\documentclass{article}
\usepackage{algorithm}
\usepackage{algpseudocode}
\usepackage{graphicx}
\usepackage{subcaption}
\usepackage{booktabs}
\usepackage[preprint]{corl_2020} 

\title{DADAgger: Disagreement-Augmented Dataset Aggregation}

\author{
  Akash Haridas\\
  \href{mailto:akashh@cs.toronto.edu}{akashh@cs.toronto.edu} 
  \And
  Karim Hamadeh\\
  \href{mailto:kar@cs.toronto.edu}{kar@cs.toronto.edu} 
  \And
  Samarendra Chandan Bindu Dash\\
  \href{mailto:dashsam1@cs.toronto.edu}{dashsam1@cs.toronto.edu} 
}

\begin{document}
\maketitle


\begin{abstract}
    \textsc{DAgger} is an imitation algorithm that aggregates its original datasets by querying the expert on all samples encountered during training. In order to reduce the number of samples queried, we propose a modification to \textsc{DAgger}, known as \textsc{DADAgger}, which only queries the expert for state-action pairs that are out of distribution (OOD). OOD states are identified by measuring the variance of the action predictions of an ensemble of models on each state, which we simulate using dropout. Testing on the Car Racing and Half Cheetah environments achieves comparable performance to \textsc{DAgger} but with reduced expert queries, and better performance than a random sampling baseline. We also show that our algorithm may be used to build efficient, well-balanced training datasets by running with no initial data and only querying the expert to resolve uncertainty.  
\end{abstract}



\section{Problem Description}

The \textsc{DAgger} algorithm \cite{DAgger} addresses the covariate shift encountered in behavioural cloning by querying the expert again on all the states that the agent encounters during a test run, then aggregating these new samples with the existing dataset. 

However, querying the expert can be costly in some situations. A drawback of \textsc{DAgger} is that it queries the expert at all collected observations without considering which actions are most valuable towards training a good policy. The DRIL algorithm \cite{dril} handles the issue of covariance shift by adding a regularizer term to the optimisation of the policy, implicitly favouring policies that choose to enter states which minimise the variance of an ensemble of policies for these states. The rationale is that the variance should be low for states that are in-distribution, so this should motivate the learner policy to stay within this distribution, assuming it is sufficient to achieve proper performance. A drawback of this method is that it may not be able to learn from incomplete datasets, and cannot explore further.

We therefore design an algorithm that does not needlessly query the expert, while also selectively querying it to learn about states which genuinely have not been encountered. This combines the flexibility and exploration of \textsc{DAgger}, while taking advantage of the ability of an ensemble to identify out-of-distribution states as done in \citet{dril}.

\section{Related Work}
Prior work in this area mainly consists of estimating uncertainty in the predictions of deep neural networks, as well as improving the sample efficiency of the \textsc{DAgger} algorithm. 

\citet{uncertainty2} learn a probability distribution on the weights of a neural network, which allows the network to make more reasonable predictions about unseen data. \citet{uncertainty1} estimate model uncertainty from the dropout layers in deep neural networks. A common way to estimate model uncertainty is with an ensemble of learners: the uncertainty is said to be high when the predictions of the ensemble have high variance. \citet{batchensemble} reduces the computational cost of ensembles by sharing some weights across the networks. \citet{dril} includes the ensemble variance in an additional loss term which implicitly trains the imitation policy to avoid states for which it has not seen demonstrations.

Several works have attempted to improve the sampling efficiency of the \textsc{DAgger} imitation learning algorithm. \citet{kim2013maximum} propose a query metric that determines how close an encountered state is from the distribution of data gathered so far, and query the expert only if this metric is above some threshold. They use Maximum Mean Discrepancy as the metric, which was originally used to determine whether two sets of data are from different probability distributions. \citet{zhang2016query} achieve a similar goal by employing a separate network, called the safety policy, to predict whether the current policy being trained is likely to make an error, and subsequently use this information to determine which visited states need to be included in the next \textsc{DAgger} iteration. Similarly, \citet{laskey2016shiv} uses a Support Vector Machine classifier to determine the risk level of a particular state. \citet{menda2017dropoutDAgger} uses ensembling to estimate model uncertainty to build a probabilistic variant of \textsc{DAgger}, where they use the model's uncertainty over its predicted actions to determine when to query the expert. They use this to enforce model safety, rather than to select when to augment the dataset in training as in our work.

\section{Methodology}
We propose a new algorithm called \textsc{DADAgger} (Disagreement-Augmented Dataset Aggregation algorithm), which is based on \textsc{DAgger}, and borrows from \cite{dril} the principle that out-of-distribution states induce higher disagreement among an ensemble of policies. We use this to modify \textsc{DAgger} to only query the expert on states with high disagreement (a particular percentage of visited states), in an attempt to gain maximal information and resolve the most uncertainty in a smaller number of queries. This improves on \textsc{DAgger} by reducing the number of expert calls, while still being able to handle incomplete datasets unlike the entirely offline DRIL. Importantly, this method does not query the expert to determine when the learner has encountered a state that it cannot handle, but rather relies entirely on the disagreement of an ensemble of learners.

Disagreement can be defined here as the variance of the output, if the outputs admit a distance between themselves, which is keeping with the implementation in \cite{dril} or possibly the entropy, if discrete and non-transitive. Since training an ensemble of neural networks is prohibitively expensive, in order to efficiently obtain the variance we instead use dropout layers to approximate uncertainty of our estimators. By passing the same input $M$ times through a single network with dropout enabled after every layer, we obtain a Gaussian Process sampling theoretically different networks \cite{dropout}. \textsc{DADAgger} is formally stated below. Tunable hyperparameters include the percentage of states to save, denoted by $\alpha$, according to their variance, and $M$, the number of policies present in our ensemble. Since we are using dropout to simulate the ensemble, the choice of $M$ does not significantly increase training time, as only one network has its weights updated. The network is called $M$ times in order to determine the uncertainty associated with a particular action. Smaller values of $\alpha$ query a smaller number of samples, but in turn make convergence more difficult due to less augmentation of the dataset. Note that during evaluation of the policy (i.e at test time), we sample the network once to predict actions, rather than obtain and ensemble mean, to ensure a fair comparison to \textsc{DAgger}.

\begin{figure}
\centering
\includegraphics[width=0.4\textwidth,height=0.21\textheight]{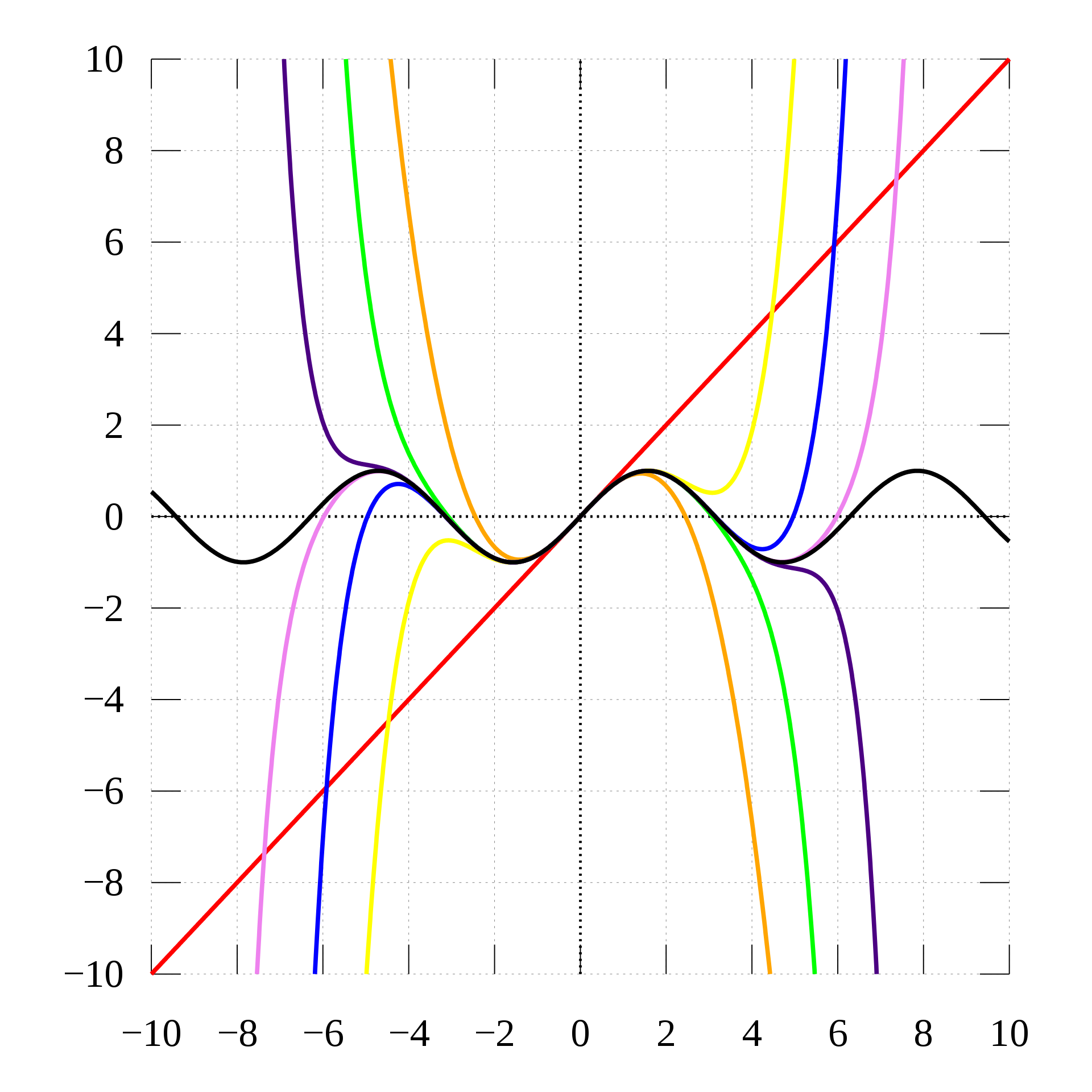}
\caption{Simplified diagram of different functions agreeing over an area of $\mathcal{R}$ and disagreeing beyond.}
\label{fig:colors}
\end{figure}

Figure \ref{fig:colors} shows a simplified schematic of different models agreeing over an area of the number line $\mathcal{R}$ centered on the origin (representing the in-distribution states), but disagreeing in areas beyond it (the out-of-distribution that is not being explicitly learned).

\begin{algorithm}
\caption{DADAgger Algorithm with Ensemble}\label{alg:cap}
\begin{algorithmic}
\State {Initialise $D$ $\gets$ $\emptyset$ }
\State {Initialise $\hat{\pi}_{1,1},\hat{\pi}_{1,2}..\hat{\pi}_{1,M}$}
\For {$i=1$ to $i=n$}
\State Sample T-step trajectories using $\hat{\pi}_{i,1}$
\State Compute Variance of $\hat{\pi}_{i,1}(s),\hat{\pi}_{i,2}(s)..\hat{\pi}_{i,M}(s)$ for all visited s.
\State Keep the $\alpha$ percent with the highest variance.
\State Get dataset $D_i={\{s,\pi^*(s)\}}$ for all filtered s.
\State Aggregate datasets $D \gets D_{i} \cup D$
\State Train classifiers $\hat{\pi}_{i+1,1},\hat{\pi}_{i+1,2}..\hat{\pi}_{i+1,M}$ on $D$
\EndFor
\State \Return best $\hat{\pi}_{i,1}$ on validation

\end{algorithmic}
\end{algorithm}
\begin{algorithm}
\caption{DADAgger Algorithm with Dropout}\label{alg:cap}
\begin{algorithmic}
\State {Initialise $D$ $\gets$ $\emptyset$ }
\State {Initialise $\hat{\pi}_1$ with dropout layers}
\For {$i=1$ to $i=n$}
\State Sample T-step trajectories using $\hat{\pi}_{i}$
\State Generate $\hat{y}_{i,j} = \hat{\pi}`_{i,j}(s)$ where $\hat{\pi}`_{i,j}$ is a variation of $\hat{\pi}_{i}$ for which a subset of neurons are turned off randomly using the dropout layer.   
\State Compute Variance of $\hat{y}_{i,1},\hat{y}_{i,2},...,\hat{y}_{i,M}$ for all visited s.
\State Keep the $\alpha$ percent with the highest variance.
\State Get dataset $D_i={\{s,\pi^*(s)\}}$ for all filtered s.
\State Aggregate datasets $D \gets D_{i} \cup D$
\State Train classifiers $\hat{\pi}_{i+1}$ on $D$
\EndFor
\State \Return best $\hat{\pi}_{i}$ on validation

\end{algorithmic}
\end{algorithm}
\section{Experiments}
\subsection{Baselines and Parameter Tuning}
We implement our method on the Car Racing environment available through OpenAI Gym \cite{gym}. For the policy network we use a CNN with dropout layers in the fully-connected section, that takes in an image of the current state and predicts a steering action.

Our first experiment consists of varying the hyper-parameters $M$ and $\alpha$ in order to gauge their effect on the convergence of \textsc{DADAgger}. Note that there are no “optimal parameters”. Increasing $M$ and $\alpha$ should hypothetically reduce the number of iterations required to reach convergence (to be verified in testing), however these increases come at the cost of training time for more models in the ensemble and more expert queries. We compare the performance and convergence properties of our algorithm to \textsc{DAgger} (obtained by setting both $\alpha$ and $M$ to 1), as well as to an agent which queries the expert at observed states selected at random (obtained by setting $M$ to 1 for varying $\alpha$) to test whether the quality of sampling plays a role in achieving convergence. We run the algorithm for 10 iterations with 5 different random seeds, and measure the proportion of runs that converged. We also measure the distribution of actions accumulated in the dataset, to obtain a quantitative assessment of the efficiency of the algorithm in exploring the action space. Reported standard deviation is that of a $p=0.5$ binomial estimator, which is guaranteed to be an upper-bound of error.

\subsection{Half Cheetah}
After implementing our algorithm on a simple environment, we test it on the Half Cheetah MuJoCo \cite{todorov2012mujoco} environment. Compared to Racer, Half Cheetah involves more complex dynamics, and a higher dimensional and continuous-valued action space. The observation is a state vector rather than an image, therefore we swap out the CNN for an MLP (with dropout layers), keeping the rest of the algorithm the same. We repeat each of the values of $M$ and $\alpha$ used in the previous experiment, and measure the Reward obtained by each policy at the end of each \textsc{DADAgger} iteration. We compare the performance and convergence properties with base \textsc{DAgger}, as well as an agent querying the expert at observed states selected at random.

\subsection{Dataset Construction}
Our final experiment exploits the efficiency of our choice of expert queries to construct a compact, and representative dataset that allows rapid, one-shot training. Instead of using an initial training dataset as in previous experiments, we run \textsc{DADAgger} with an empty initial dataset, and consequently create the dataset over several iterations, querying the expert only to resolve uncertainty. Our hypothesis is that the final dataset upon convergence will be fairly small, due to the inclusion of only points which the models disagree on at different stages of training. We use $\alpha=0.1$, as well as $50$ iterations, which is the time we found needed to obtain a proper convergence using this technique. We then measure the distribution of actions in our final dataset and compare it to the dataset produced by \textsc{DAgger} and \textsc{DADAgger} when supplied with an initial dataset.

\section{Results and Discussion}
\subsection{Baselines and Parameter Tuning}
The results are shown in Table~\ref{table:racecar_results}. For $\alpha=0.4$, all trained models converge in all runs, including random querying. We hypothesise that this is due to the relatively small difference between consecutive frames, which means skipping a frame, on average, will not be overly detrimental to training. With a lower $\alpha$ of $0.2$, our algorithm, regardless of the choice of $M$, maintains 100\% convergence, while random querying only succeeds 40\% of the time. Our algorithm thus outperforms the random baseline on this metric ($p<0.003$). Finally, on the lowest setting of $\alpha$, random sampling fails to converge entirely, while our algorithm only sometimes converges. This convergence often happens in the 9th or 10th iteration, so it is entirely possible that convergence could follow in the cases where it did not terminate, however we are restricting testing to 10 iterations to control across all methods. Note that the difference between $M=25$ and $M=10$ is not statistically significant.

To verify the mechanism by which our improvement is obtained, we obtain the histogram of saved actions in different settings in Figure~\ref{fig:histograms} . \textsc{DAgger} is seen to heavily sample straight-line actions (close to the middle). Our algorithms are seen to substantially increase the proportion of sampled turning actions with respect to the size of the dataset, but also use less sampling overall. Interestingly, the starting dataset is more balanced than any of the ones resulting from \textsc{DAgger} / \textsc{DADAgger} iterations, yet fails to converge, meaning the quality of a dataset is not entirely determined by how balanced it is.

\subsection{Half Cheetah}
\begin{figure}
\centering
\includegraphics[width=0.8\textwidth]{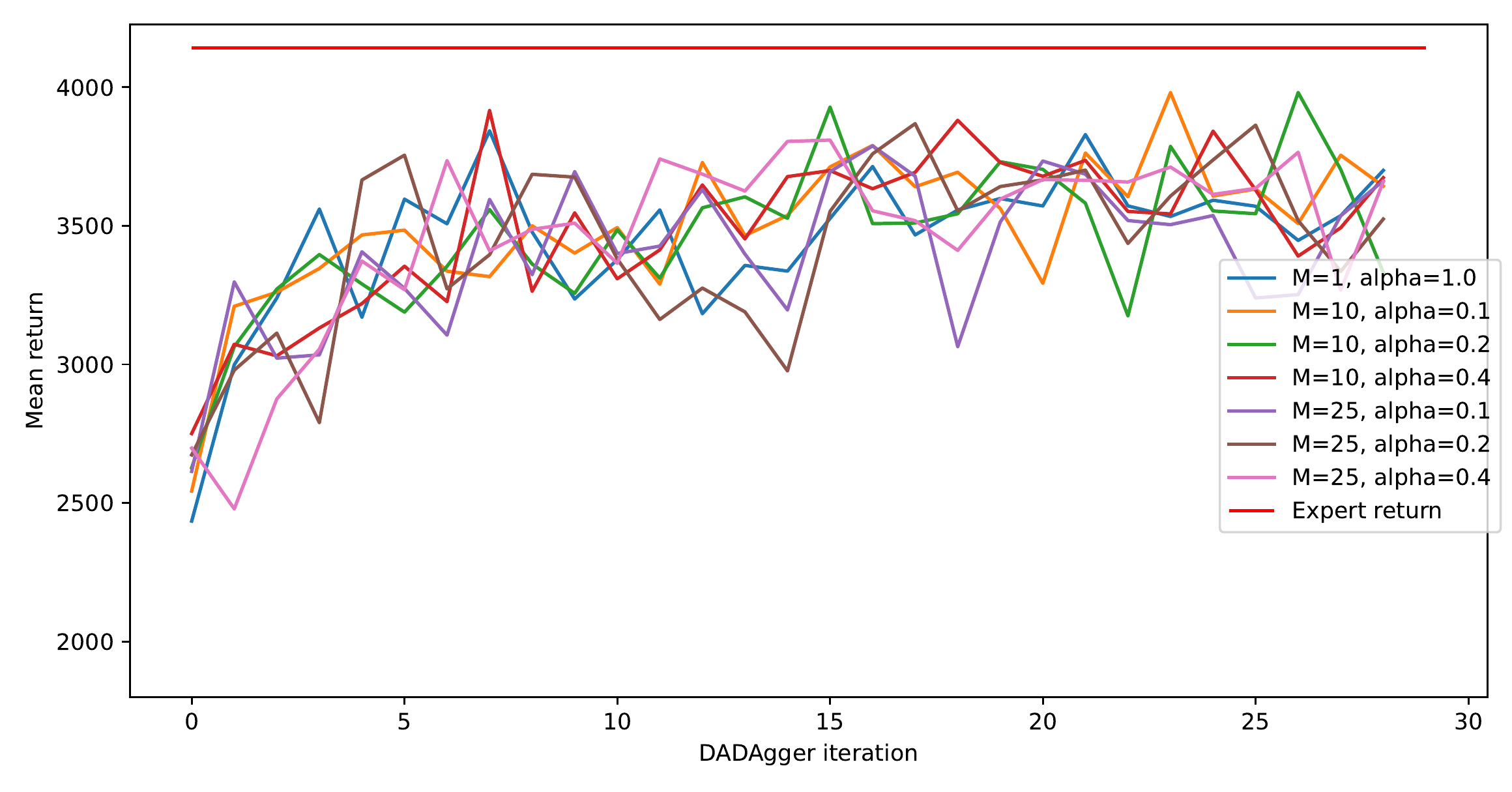}
\caption{All variants of \textsc{DADAgger} converge to a similar performance as base \textsc{DAgger} ($M = 1$ and $\alpha = 1$) on the Half Cheetah environment, with significantly less queries to the expert. However, none of the methods achieve a performance comparable to the expert}
\label{fig:halfcheetah_convergence}
\end{figure}

Figure~\ref{fig:halfcheetah_convergence} shows that all variants of \textsc{DADAgger} converge to a similar performance as base \textsc{DAgger} on Half Cheetah, which demonstrates that our uncertainty measurement and sampling technique is also capable of handling a multi-dimensional and continuous action space. It achieves this with as little as 10\% of queries to the expert ($\alpha=0.1$) as base \textsc{DAgger}, again suggesting that it samples only the most important datapoints to learn a policy and resolve uncertainty. Interestingly, neither of the methods (incl. base \textsc{DAgger}) matched the expert's performance, possibly due to the increased complexity of the environment.

\subsection{Dataset Construction}
\begin{figure}[h]
     \centering
     \begin{subfigure}[b]{0.4\textwidth}
         \centering
         \includegraphics[width=\textwidth]{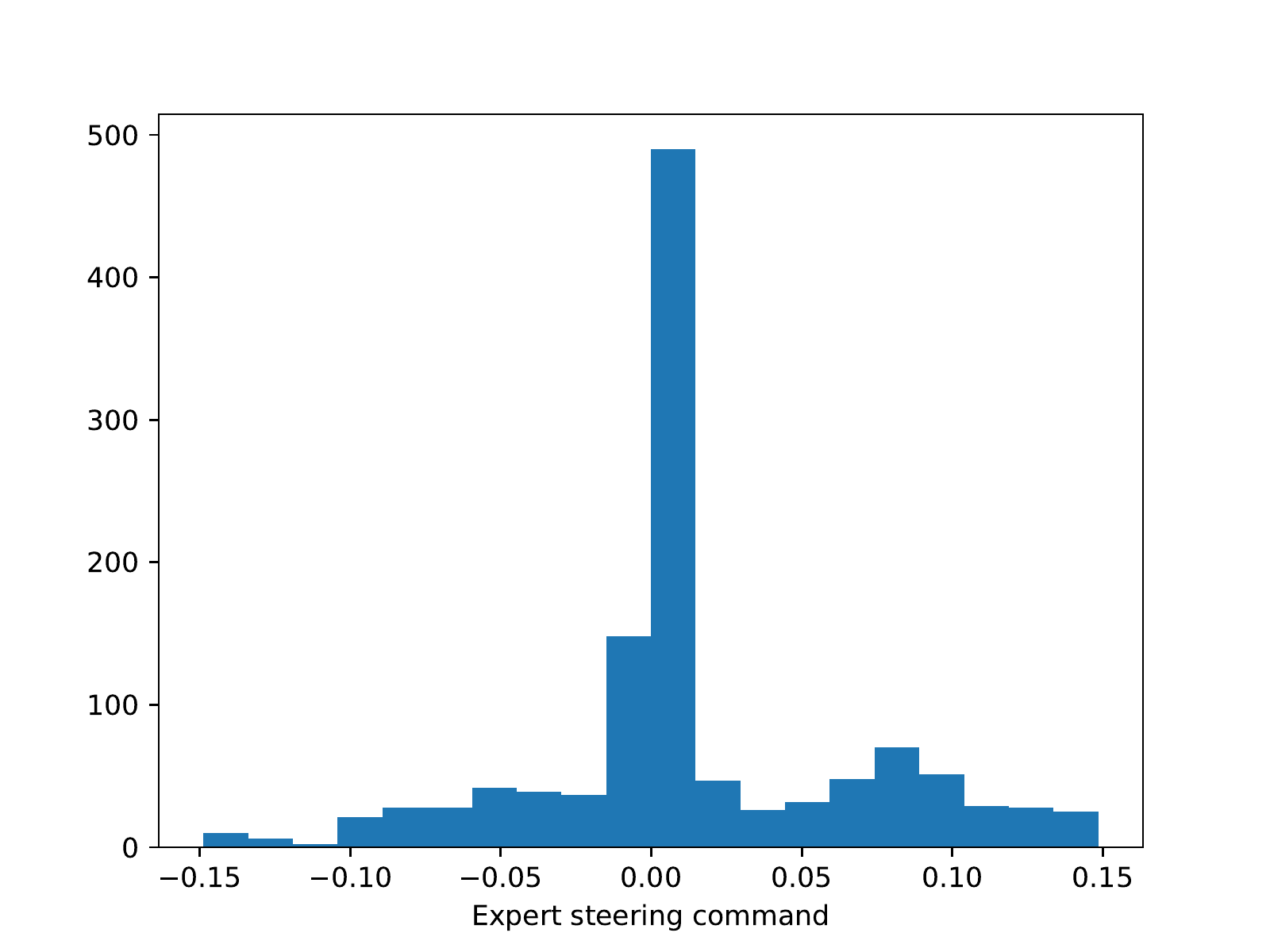}
         \caption{Starting dataset}
         \label{fig:hist_basedataset}
     \end{subfigure}
     \begin{subfigure}[b]{0.4\textwidth}
         \centering
         \includegraphics[width=\textwidth]{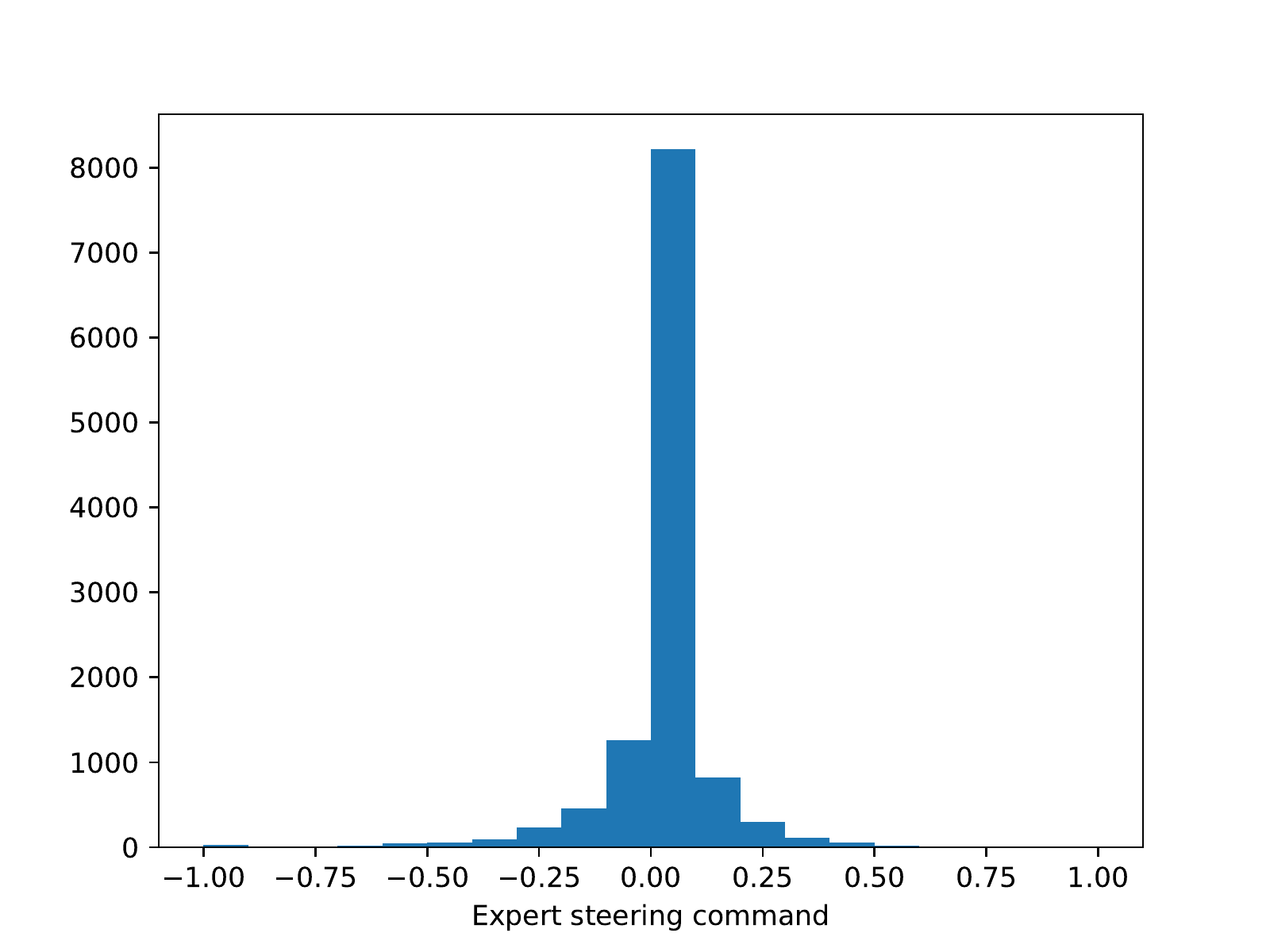}
         \caption{Dataset after a \textsc{DAgger} run}
         \label{fig:hist_basedagger}
     \end{subfigure}
     \begin{subfigure}[b]{0.4\textwidth}
         \centering
         \includegraphics[width=\textwidth]{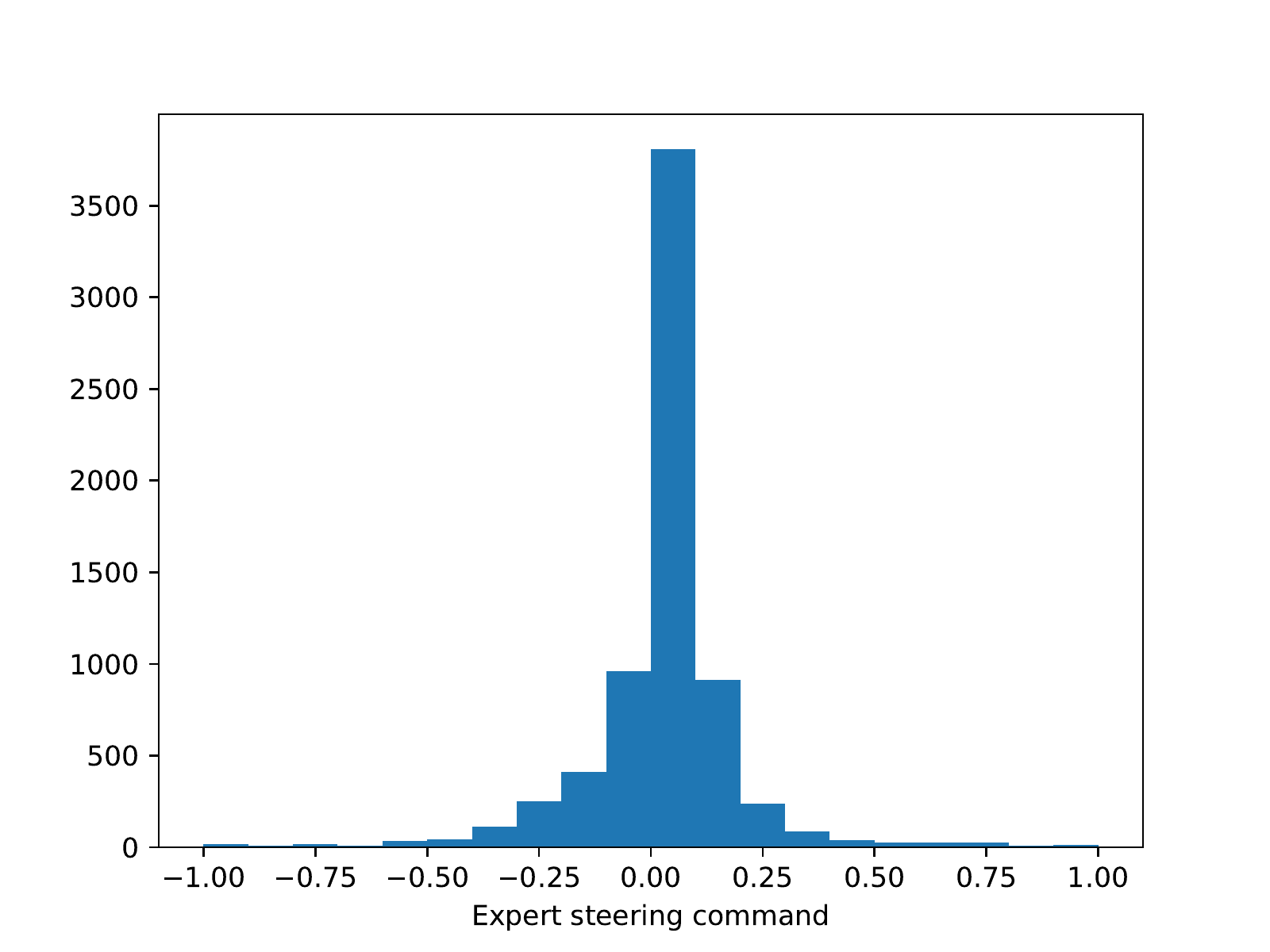}
         \caption{\textbf{Ours}: Dataset after a \textsc{DADAgger} run with $M=10$, $\alpha=0.4$}
         \label{fig:hist_m10alpha4}
     \end{subfigure}
     \begin{subfigure}[b]{0.4\textwidth}
         \centering
         \includegraphics[width=\textwidth]{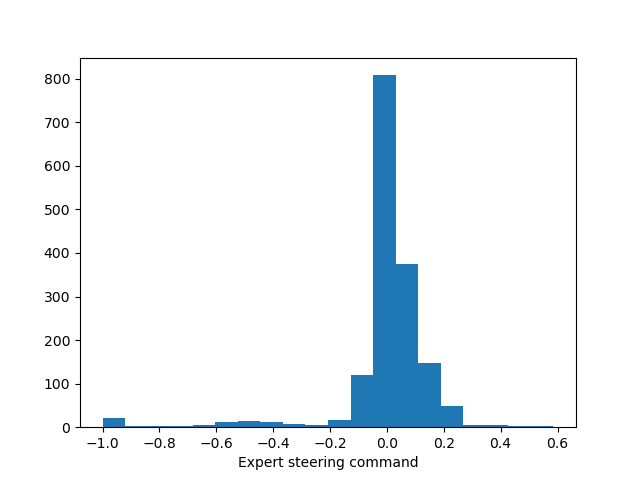}
         \caption{\textbf{Ours}: Dataset after a \textsc{DADAgger} run with $M=10$, $\alpha=0.1$}
         \label{fig:hist_m10alpha1}
     \end{subfigure}
     \begin{subfigure}[b]{0.4\textwidth}
         \centering
         \includegraphics[width=\textwidth]{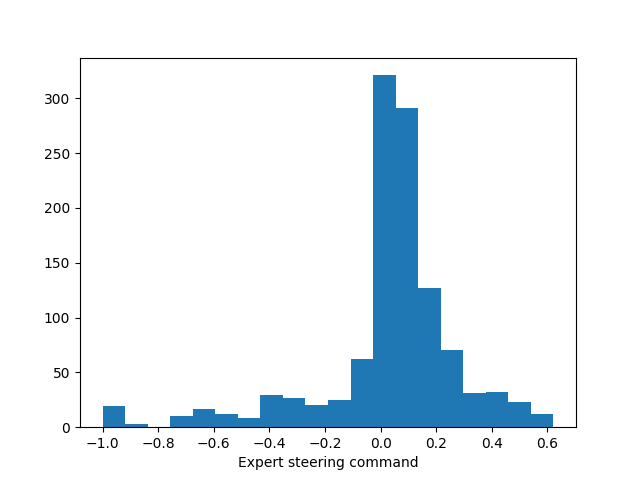}
         \caption{\textbf{Ours}: Efficient dataset after a \textsc{DADAgger} run starting with an empty intial dataset.}
         \label{fig:hist_null}
     \end{subfigure}
        \caption{Comparison of sampling distributions on the Car Racing environment. While the initial dataset mostly contains straight-line examples, \textsc{DAgger} samples at least some points at turns. Our algorithm DADagger overall samples far fewer examples by being less wasteful with sampling at straight lines (notice the difference in scale in the y-axes). Starting \textsc{DADAgger} with a null initial datset produces the most balanced of all datasets, especially with respect to extreme sharp turns which are relatively undersampled in other experiments.}
        \label{fig:histograms}
\end{figure}

Convergence is obtained at roughly iteration 50 (hence our decision to stop). We are more interested in examining the nature of the constructed dataset over the performance of the policy. The final dataset consists of 746 samples, which is markedly less than even the initial dataset we were using for \textsc{DAgger} and \textsc{DADAgger}, which contained 1139 samples. It is thus far smaller than datasets generated by the previous experiments, regardless of the choice of $\alpha$. This is notable because a policy trained on this dataset is able to converge with 0 augmentation (ie, 1 \textsc{DAgger} iteration, or in one-shot), indicating a high quality of samples. We once again look to the histogram (Figure~\ref{fig:hist_null}), and find it to be the most balanced of all datasets, especially with respect to extreme sharp turns which were relatively undersampled even in \textsc{DADAgger} experiments. \textsc{DADAgger} thus acts as a way to create small and efficient datasets for one-shot training.

\begin{table}[]
\centering
\begin{tabular}{@{}lllll@{}}
\toprule
$M$ / $\alpha$ & 0.1              & 0.2 & 0.4 \\ \midrule
10             & $0\pm 20$ &    $100\pm 20$  & $100\pm 20$ \\
25             & $60\pm 20$ &    $100\pm 20$   & $100\pm 20$ \\
Random         & $40\pm 20$               & $40\pm 20$    & $100\pm 20$ \\ \toprule
\end{tabular}
\caption{Percentage of trials that led to a successful lap on the Car Racing environment.}
\label{table:racecar_results}
\end{table}

\section{Limitations and Conclusion}
While \textsc{DADAgger} is able to improve on the sampling efficiency of \textsc{DAgger}, a few weaknesses arise. Primarily, the agent is still required to engage in the environment as many times as \textsc{DAgger}, meaning there is no improvement on the number of episodes required to converge, should that be expensive. This is particularly true in the case where the dataset was initialised to zero, as many more episodes were required to achieve convergence. Another weakness we occasionally observed in our tests was high confidence predictions for wrong outcomes, which could lead to skipping essential sampling and repeating the same error. It is therefore absolutely essential for our algorithm to function, that a certain degree of independence exist between predictors. Any bias will create agreement in out-of-distribution states that may lead to the algorithm getting stuck in a cycle of not sampling, failing because of that error, and not sampling the requisite states again. Indeed, 30 of 50 iterations in the dataset construction experiment were failures of relatively similar form, which could hint at the existence of this problem in sparse datasets.

In conclusion, this paper introduces \textsc{DADAgger}, a novel method to increase querying efficiency for the \textsc{DAgger} algorithm without compromising convergence. Our method matches \textsc{DAgger} performance on both the Car Racing and Half Cheetah environments. Furthermore, our technique is efficient in its execution due to the use of dropout, and could also find use as a dataset generator due to its selectivity. Further avenues of exploration involve exploration of alternate methods to assess uncertainty that could allow the use of an absolute metric rather than the fraction $\alpha$, and would not require several network evaluations for each action. Another uncovered area is treating the sample space without regard to the action, and attempting to devise a measure of how out-of-distribution a particular sample is that is completely independent of the predictor which generates the action.

\section*{Contributions}
KH was responsible for the algorithm design and Experiment 3. Experiment 1 was split among all members. AH and SCBD conducted Experiment 2. SCBD and AH jointly wrote/adapted and optimized most of the code used to perform all experiments.


\bibliography{example}  

\begin{thebibliography}{12}
\providecommand{\natexlab}[1]{#1}
\providecommand{\url}[1]{\texttt{#1}}
\expandafter\ifx\csname urlstyle\endcsname\relax
  \providecommand{\doi}[1]{doi: #1}\else
  \providecommand{\doi}{doi: \begingroup \urlstyle{rm}\Url}\fi

\bibitem[Ross et~al.(2011)Ross, Gordon, and Bagnell]{DAgger}
S.~Ross, G.~Gordon, and D.~Bagnell.
\newblock A reduction of imitation learning and structured prediction to
  no-regret online learning.
\newblock In \emph{Proceedings of the fourteenth international conference on
  artificial intelligence and statistics}, pages 627--635. JMLR Workshop and
  Conference Proceedings, 2011.

\bibitem[Brantley et~al.(2019)Brantley, Sun, and Henaff]{dril}
K.~Brantley, W.~Sun, and M.~Henaff.
\newblock Disagreement-regularized imitation learning.
\newblock In \emph{International Conference on Learning Representations}, 2019.

\bibitem[Blundell et~al.(2015)Blundell, Cornebise, Kavukcuoglu, and
  Wierstra]{uncertainty2}
C.~Blundell, J.~Cornebise, K.~Kavukcuoglu, and D.~Wierstra.
\newblock Weight uncertainty in neural network.
\newblock In \emph{International conference on machine learning}, pages
  1613--1622. PMLR, 2015.

\bibitem[Gal and Ghahramani(2016)]{uncertainty1}
Y.~Gal and Z.~Ghahramani.
\newblock Dropout as a bayesian approximation: Representing model uncertainty
  in deep learning.
\newblock In \emph{international conference on machine learning}, pages
  1050--1059. PMLR, 2016.

\bibitem[Wen et~al.(2020)Wen, Tran, and Ba]{batchensemble}
Y.~Wen, D.~Tran, and J.~Ba.
\newblock Batchensemble: an alternative approach to efficient ensemble and
  lifelong learning.
\newblock \emph{arXiv preprint arXiv:2002.06715}, 2020.

\bibitem[Kim and Pineau(2013)]{kim2013maximum}
B.~Kim and J.~Pineau.
\newblock Maximum mean discrepancy imitation learning.
\newblock In \emph{Robotics: Science and systems}, 2013.

\bibitem[Zhang and Cho(2016)]{zhang2016query}
J.~Zhang and K.~Cho.
\newblock Query-efficient imitation learning for end-to-end autonomous driving.
\newblock \emph{arXiv preprint arXiv:1605.06450}, 2016.

\bibitem[Laskey et~al.(2016)Laskey, Staszak, Hsieh, Mahler, Pokorny, Dragan,
  and Goldberg]{laskey2016shiv}
M.~Laskey, S.~Staszak, W.~Y.-S. Hsieh, J.~Mahler, F.~T. Pokorny, A.~D. Dragan,
  and K.~Goldberg.
\newblock Shiv: Reducing supervisor burden in dagger using support vectors for
  efficient learning from demonstrations in high dimensional state spaces.
\newblock In \emph{2016 IEEE International Conference on Robotics and
  Automation (ICRA)}, pages 462--469. IEEE, 2016.

\bibitem[Menda et~al.(2017)Menda, Driggs-Campbell, and
  Kochenderfer]{menda2017dropoutDAgger}
K.~Menda, K.~Driggs-Campbell, and M.~J. Kochenderfer.
\newblock Dropoutdagger: A bayesian approach to safe imitation learning.
\newblock \emph{arXiv preprint arXiv:1709.06166}, 2017.

\bibitem[Srivastava et~al.(2014)Srivastava, Hinton, Krizhevsky, Sutskever, and
  Salakhutdinov]{dropout}
N.~Srivastava, G.~Hinton, A.~Krizhevsky, I.~Sutskever, and R.~Salakhutdinov.
\newblock Dropout: a simple way to prevent neural networks from overfitting.
\newblock \emph{The journal of machine learning research}, 15\penalty0
  (1):\penalty0 1929--1958, 2014.

\bibitem[Brockman et~al.(2016)Brockman, Cheung, Pettersson, Schneider,
  Schulman, Tang, and Zaremba]{gym}
G.~Brockman, V.~Cheung, L.~Pettersson, J.~Schneider, J.~Schulman, J.~Tang, and
  W.~Zaremba.
\newblock Openai gym, 2016.

\bibitem[Todorov et~al.(2012)Todorov, Erez, and Tassa]{todorov2012mujoco}
E.~Todorov, T.~Erez, and Y.~Tassa.
\newblock Mujoco: A physics engine for model-based control.
\newblock In \emph{2012 IEEE/RSJ international conference on intelligent robots
  and systems}, pages 5026--5033. IEEE, 2012.

\end{thebibliography}

\end{document}